\documentclass{article}

\PassOptionsToPackage{numbers, compress}{natbib}

\usepackage[final]{neurips_2022} 

\usepackage[utf8]{inputenc} 
\usepackage[T1]{fontenc}    
\usepackage{hyperref}       
\usepackage{url}            
\usepackage{booktabs}       
\usepackage{amsfonts}       
\usepackage{nicefrac}       
\usepackage{microtype}      
\usepackage{xcolor}         
\usepackage{color}
\usepackage{amsmath}
\usepackage{amssymb}
\usepackage{subfigure}
\usepackage{booktabs} 
\usepackage{graphicx}
\usepackage{algorithmic}

\usepackage{natbib}
\setcitestyle{nonatbib}
\setcitestyle{square}

\usepackage{graphicx}
\usepackage{amssymb}
\usepackage{multirow}
\usepackage{booktabs} 
\usepackage{diagbox}
\usepackage{multicol} 
\usepackage{arydshln}
\usepackage{dashrule}
\usepackage{paralist}
\usepackage{wrapfig,lipsum,booktabs}
\usepackage{enumitem}
\usepackage{comment}
\usepackage{amsmath,amssymb} 

\usepackage[linesnumbered,ruled,lined]{algorithm2e}
\usepackage{algorithmic}

\newsavebox\CBox
\def\textBF#1{\sbox\CBox{#1}\resizebox{\wd\CBox}{\ht\CBox}{\textbf{#1}}}

\title{Task-Free Continual Learning via \\
Online Discrepancy Distance Learning}

\author{%
  Fei Ye and Adrian G. Bors  \vspace*{0.2cm} \\
  Department of Computer Science\\
  University of York\\
  York, YO10 5GH, UK \\
  \texttt{\{fy689,adrian.bors\}@york.ac.uk} \\
}

\begin{document}

\setlength{\abovedisplayskip}{5pt} 
\setlength{\belowdisplayskip}{5pt}
\setlength{\abovecaptionskip}{5pt}

\maketitle

\begin{abstract}
Learning from non-stationary data streams, also called Task-Free Continual Learning (TFCL) remains challenging due to the absence of explicit task information. Although recently some methods have been proposed for TFCL, they lack theoretical guarantees. Moreover, forgetting analysis during TFCL was not studied theoretically before. This paper develops a new theoretical analysis framework which provides generalization bounds based on the discrepancy distance between the visited samples and the entire information made available for training the model. This analysis gives new insights into the forgetting behaviour in classification tasks. Inspired by this theoretical model, we propose a new approach enabled by the dynamic component expansion mechanism for a mixture model, namely the Online Discrepancy Distance Learning (ODDL). ODDL estimates the discrepancy between the probabilistic representation of the current memory buffer and the already accumulated knowledge and uses it as the expansion signal to ensure a compact network architecture with optimal performance. We then propose a new sample selection approach that selectively stores the most relevant samples into the memory buffer through the discrepancy-based measure, further improving the performance. We perform several TFCL experiments with the proposed methodology, which demonstrate that the proposed approach achieves the state of the art performance.
\end{abstract}

\section{Introduction}

Continual learning (CL) and its extension to lifelong learning, represents one of the most desired functions in an artificial intelligence system, representing the capability of learning new concepts while preserving the knowledge of past experiences \cite{LifeLong_review}. Such an ability can be used in many real-time applications such as robotics, health investigative systems, autonomous vehicles \cite{AutonomousCar} or for guiding agents exploring artificial (meta) universes, requiring adapting to a changing environment. Unfortunately, modern deep learning models suffer from a degenerated performance on past data after learning novel knowledge, a phenomenon called catastrophic forgetting \cite{LessForgetting}. 

A popular attempt to relieve forgetting in CL is by employing a small memory buffer to preserve a few past samples and replay them when training on a new task \cite{GradientLifelong,TinyLifelong}. However, when there are restrictions on the available memory capacity, memory-based approaches would suffer from degenerated performance on past tasks, especially when aiming to learn an infinite number of tasks. Recently, the Dynamic Expansion Model (DEM) \cite{LifelongInfinite} has shown promising results in CL, aiming to guarantee optimal performance by preserving the previously learnt knowledge through the parameters of frozen components trained on past data, while adding a new component when learning a novel task. However, such approaches require knowing where and when the knowledge associated with a given task is changed, which is not always applicable in a real environment.

In this paper, we address a more realistic scenario, called Task-Free Continual Learning (TFCL) \cite{taskFree_CL}, where task identities are not available while the model can only access a small batch of samples at a given time. Most existing CL methods requiring the task label can be adapted to TFCL by removing the task information dependency. For instance, memory-based approaches can store a few past samples from the data stream at each training time and replay them during later training steps \cite{ContinualPrototype,GradientTFCL}. However, such an approach requires to carefully design the sample selection criterion to avoid memory overload. The key challenge for the memory-based approach is the negative backward transfer caused by the stored samples that interfere with the model's updating with incoming samples  \cite{TinyLifelong}. This issue can be relieved by DEM in which previously learnt samples are preserved into frozen components and do not interfere with the learning of probabilistic representations of new data \cite{,NeuralDirichelt,LifelongUnsupervisedVAE}. However, these approaches do not provide any theoretical guarantees and there are no studies analysing the trade-off between the model's generalization and its complexity under TFCL. 

Recent attempts have provided the theoretical analysis for CL from different perspectives including the risk bound \cite{LifelongVAEGAN,LifelongInfinite}, NP-hard problem \cite{OptimalCL}, Teacher-Student framework \cite{CLTeacherStudent,LifelongTeacherStudent} and game theory \cite{CL_TradeOff}. However all these approaches require strong assumptions, such as defining the task identities, which is not available in TFCL. This inspires us to bridge the gaps between the underlying theory and the algorithm implementation for TFCL. We propose a theoretical classification framework, which provides new insights in the forgetting behaviour analysis and guidance for algorithm design addressing catastrophic forgetting. The primary motivation behind the proposed theoretical framework is that we can formulate forgetting as a generalization error in the domain adaptation theory. Based on this analysis we extend the domain adaptation theory \cite{domainTheory} to derive time-dependent generalization risk bounds, explicitly explaining the forgetting process at each training step.

Inspired by the theory, we devise the Online Discrepancy Distance Learning (ODDL) method which introduces a new expansion mechanism based on the discrepancy distance estimation for implementing TFCL. The proposed expansion mechanism detects the data distribution shift by evaluating the variance of the discrepancy distance during the training. This model enables a trade-off mechanism between the model's generalization and complexity. We also propose a new sample selection approach based on the discrepancy-based criterion, which guides storing diverse samples with respect to the already learnt knowledge, further improving performance. Our contributions are~: 
\begin{itemize}[leftmargin=10pt]
\setlength{\itemsep}{1pt}
\setlength{\parsep}{1pt}
\setlength{\parskip}{1pt}
\item This paper is the first research study to propose a new theoretical framework for TFCL, which provides new insights into the forgetting behaviour of the model in classification tasks.
\item Inspired by the theoretical analysis, we develop a novel dynamic expansion approach, which ensures a compact model architecture enabled by optimal performance.
\item We propose a new sample selection approach that selects appropriate data samples for the memory buffer, further improving performance.
\item The proposed method achieves state of the art results on TFCL benchmarks,
\end{itemize} 

\section{Related works}

\noindent \textBF{Continual learning} defines a learning paradigm which aims to learn a sequence of tasks without forgetting. Catastrophic forgetting is a major challenge in continual learning. One of the most popular approaches to relieve forgetting is by imposing a regularization loss within the optimization procedure \cite{BoostingTransfer,Distilling_nets,LessForgetting,EWC,CL_Bayesian,Lwf,TRGP,VCL,LearnAdd,LookingBack,LifeLong_combination,OnlineStructuredLaplace,LifelongGraph}, where the network's parameters which are important to the past tasks are penalized when updating. Another kind of approaches for continual learning focuses on the memory system, which usually employs a small memory buffer \cite{GradientLifelong,RainbowMemory,TinyLifelong,RepresentationalContinuity,Gdumb,InfomrationCL,OnlineCoreset} to store a few past data or trains a generator to provide the replay samples when learning new tasks \cite{LifelongUnsupervisedVAE,Generative_replay,LifelongVAEGAN,Lifelonginterpretable,LifelongMixuteOfVAEs,LifelongTeacherStudent,LifelongTwin}. However, these approaches usually rely on knowing the task information, which is not applicable in TFCL.

\noindent \textBF{Task-free continual learning} is a special scenario in CL where a model can only see one or very few samples in each training step/time without having any task labels. Using a small memory buffer to store past samples has shown benefits for TFCL and was firstly investigated in \cite{taskFree_CL,OCM,LifelongEvolved}. This memory replay approach was then extended by employing Generative Replay Mechanisms (GRMs) for training both a Variational Autoencoder (VAEs) \cite{VAE} and a classifier, where a new retrieving mechanism is used to select specific data samples, called the Maximal Interfered Retrieval (MIR), \cite{OnlineContinualLearning}. The Gradient Sample Selection (GSS) \cite{GradientLifelong} is another sample selection approach that treats sample selection as a constrained optimization reduction. More recently, a Learner-Evaluator framework is proposed for TFCL, called the Continual Prototype Evolution (CoPE) \cite{ContinualPrototype} which stores the same number of samples for each class in the memory in order to ensure the balance replay. Another direction for the memory-based approaches is to edit the stored samples which would increase the loss in the upcoming model updates, called the Gradient based Memory EDiting (GMED) \cite{GradientTFCL}, which can be employed in the existing CL models to further enhance their performance.

\noindent \textBF{Dynamic expansion models} aim to automatically increase the model's capacity to adapt to new tasks by adding new hidden layers and units. Such an approach, called the Continual Unsupervised Representation Learning (CURL) \cite{LifelongUnsupervisedVAE},  dynamically builds new inference models when meeting the expansion criterion. However, since CURL still requires a GRM to relieve forgetting, it would lead to a negative knowledge transfer when updating the network's parameters to adapt to a new task. This issue can be addressed by using Dirichlet processes by adding new components while freezing all previously learnt members, in a model called the Continual Neural Dirichlet Process Mixture (CN-DPM), \cite{NeuralDirichelt}. However, the expansion criterion used by these approaches relies on the change of the loss when training each time, which does not have any theoretical guarantees.

\section{Theoretical analysis of TFCL}

In this section, we firstly introduce learning settings and notations, and then we analyze the forgetting behaviour for a single as well as for a dynamic expansion model by deriving their Generalization Bounds (GBs).

\subsection{Preliminary}

Let $\mathcal{X}$ be the input space and $\mathcal{Y}$\vspace{-1.0pt} represent the output space which is $\{-1,1\}$ for binary classification and $\{1,2,\dots,n'\},n'>2$ for multi-class classification.\vspace{-2.5pt} Let $\mathcal{D}^T_i = \{{\bf x}^T_j,y^T_j \}^{N^T_i}_{j=1}$ and $\mathcal{D}^S_i = \{{\bf x}^S_j,y^S_j \}^{N^S_i}_{j=1}$ represent the training and testing sets for the $i$-th dataset where ${\bf x}^T_j \in \mathcal{X}$ and $y^T_j \in {\mathcal{Y}}$ are the image and the associated ground truth label. $N^T_i$ and $N^S_i$ are the total number of samples for ${\mathcal{D}}^T_i$ and ${\mathcal{D}}^S_i$, respectively. In the paper, we mainly focus on the task-free class-incremental learning, described as follows.

\noindent \textBF{Definition 1.} \textBF{(Data stream.)} For a given $t$-th training dataset $\mathcal{D}^S_t$ with $C^S_t$ data categories, let us consider a data stream $\mathcal{S}$ which consists of samples 
$\mathcal{D}^S_{t,j}$ from\vspace{-2.5pt} each category,\vspace{-1.0pt} expressed by $S =  \bigcup_{j=1}^{C^S_t} \mathcal{D}^S_{t,j}$.\vspace{-1.5pt} Let ${\mathcal{D}}^T_{t,j}$ represent the set of samples drawn from the $j$-th category of ${\mathcal{D}}^T_{t}$. Let ${\mathbb{P}}^T_{t,j}$ and ${\mathbb{P}}^S_{t,j}$ represent the distributions for ${\mathcal{D}}^T_{t,j}$ and ${\mathcal{D}}^S_{t,j}$, respectively. Let ${\mathbb{P}}^{T,{\mathcal{X}} }_{t,j}$ represent the marginal distribution over $\mathcal{X}$.

\noindent \textBF{Definition 2.} \textBF{ (Learning setting.)} Let $\mathcal{T}_i$ represent the $i$-th training step. For a given data stream $\mathcal{S}$, we assume that there are a total of $n$ training steps for $\mathcal{S}$, where each training step $\mathcal{T}_i$ is associated with a small batch of paired samples $\{ {\bf X}^b_i, {\bf Y}^b_i\}$ drawn from $\mathcal{S}$, expressed by ${\mathcal{S}} = \bigcup_{i=1}^{n} \{ {\bf X}_i^b, {\bf Y}^b_i \}, \;\; \{ {\bf X}_i^b, {\bf Y}^b_i \} \cap \{ {\bf X}_j^b, {\bf Y}^b_j \} = \varnothing$, where $i \ne j$ and a model (classifier) can only access $\{ {\bf X}^b_i, {\bf Y}^b_i\}$ at $\mathcal{T}_i$ while all previous batches are not available. After finishing all training steps, we evaluate the classification accuracy of the model on the testing set $\mathcal{D}^T_t$. In the following, we define the model and the memory buffer.

\noindent \textBF{Definition 3.} \textBF{(Model and memory.)} Let us consider $h$ a model implemented by a classifier, and $\mathcal{H} =\{ h \,|\, h \colon {\mathcal{X}} \to \mathcal{Y} \}$ the space of classifiers. Let $\mathcal{M}_i$ be a memory buffer at $\mathcal{T}_i$. We assume that $\mathcal{M}_i$ randomly removes samples from the memory buffer while continually adding new data to its memory at each training step. Let $\mathbb{P}_{{\mathcal{M}}_i}$ represent the probabilistic representation of ${\mathcal{M}}_i$ and $|{\mathcal{M}}_i|$ its cardinality.

\subsection{Measuring the distribution shift}
\label{distributionShift}

In TFCL, the distance between the target domain (testing set) and the source domain (memory) would be dynamically changed during each training step. We can use the discrepancy distance \cite{domainTheory} to measure this gap through the analysis of the model's risk.  

\noindent \textBF{Definition 4.} \textBF{(The risk.)} For a given distribution ${\mathbb{P}}^S_{t,j}$, the risk of a model $h$ is defined as ${\mathcal{R}} \big(h,{\mathbb{P}}^S_{t,j} \big)  \buildrel \Delta \over =  {\mathbb{E}}_{\{ {\bf x},y \} \sim {\mathbb{P}}^S_{t,j} } \big[ {\mathcal{L}} \big( y,h({\bf x}) \big) \big]$ where $\mathcal{L} \colon {\mathcal{Y}} \times {\mathcal{Y}} \to [0, 1]$ is the loss function.

\noindent \textBF{Definition 5.} \textBF{(Discrepancy distance.)} For two given distributions ${{\mathbb{P}} }^{T}_{t,j}$ and ${{\mathbb{P}}}^S_{t,j}$, the discrepancy distance $\mathcal{L}_d$ is defined on two marginals as~:
\begin{align}
{\mathcal{L}}_d \big({\mathbb{P} }^{T,{\mathcal{X}}}_{t,j}, {{\mathbb{P}}}^{S,{\mathcal{X}}}_{t,j}  \big) & \buildrel \Delta \over =   \mathop {\sup }\limits_{\left( h,h' \right) \in \mathcal{H}^2} \Big| {\mathbb{E}}_{{\bf x} \sim {\mathbb{P} }^{T,{\mathcal{X}}}_{t,j} }  \big[ {\mathcal{L}} \big( h({\bf x}),h'({\bf x}) \big)   \big]  - 
{\mathbb{E}}_{{\bf x} \sim {\mathbb{P} }^{S,{\mathcal{X}}}_{t,j} }  \big[ {\mathcal{L}} \big( h({\bf x}),h'({\bf x}) \big)   \big]
\Big|\,,
\end{align}
where $\{h,h' \} \in \mathcal{H}$. In practice, the discrepancy distance ${\mathcal{L}}_d(\cdot,\cdot)$ can be estimated as the upper bound based on the Rademacher complexity which is used in the domain adaptation theory as a measure of richness for a particular hypothesis space \cite{Theory_drifiting,BridgingTheory}.

\noindent \textBF{Corollary 1.} \cite{domainTheory}
\label{corollary1}
For two given domains\vspace{-2.0pt} ${{\mathbb{P}} }^{T,{\mathcal{X}}}_{t,j}$ and ${{\mathbb{P}}}^{S,{\mathcal{X}}}_{t,j}$, let $U_{\mathcal{P}}$ and $U_{\mathbb{P}}$ represent sample sets of sizes $m_{\mathcal{P}}$ and $m_{\mathbb{P}}$, drawn independently from ${{\mathbb{P}} }^{T,{\mathcal{X}}}_{t,j}$ and ${{\mathbb{P}}}^{S,{\mathcal{X}}}_{t,j}$.\vspace{-1.0pt} Let ${\widehat{\mathbb{P}}}^{T, {\mathcal{X}}}_{t,j}$ and ${\widehat{\mathbb{P}}}^{S,{\mathcal{X}}}_{t,j}$ represent the empirical distributions for $U_{\mathcal{P}}$ and $U_{\mathbb{P}}$. Let $\mathcal{L}({\bf x'},{\bf x}) = |{\bf x'} - {\bf x}{|^1}$ be a loss function (L1-Norm), satisfying $\forall ({\bf{x}},{\bf{x'}}) \in \mathcal{X},\mathcal{L}({\bf{x}},{\bf{x'}}) > M$, where $M>0$. Then, with the probability $1-\delta$, we have~:
\begin{align}
\label{colorllary_eq1}
    {\mathcal{L}_d}({{\mathbb{{ P}}} }^{T, {\mathcal{X}}}_{t,j}, {{\mathbb{{ P}}}}^{S, {\mathcal{X}}}_{t,j})
 &\le {\mathcal{L}_d}(\widehat{\mathbb{P} }^{T,{\mathcal{X}}}_{t,j}, \widehat{\mathbb{P}}^{S, {\mathcal{X}}}_{t,j})  + C^\star  + 3M\left( {\sqrt {\frac{{\log \left( {\frac{4}{\delta }} \right)}}{{2{m_{\mathcal{P}}}}}}  + \sqrt {\frac{{\log \left( {\frac{4}{\delta }} \right)}}{{2{m_{\mathbb P}}}}} } \right) , 
\end{align}
\noindent where $C^\star = 4q\left( {{{\mathop{\rm Re}\nolimits} }_{{U_{\mathcal P}}}}\left(\mathcal{H} \right) 
+
{{{\mathop{\rm Re}\nolimits} }_{{U_{\mathbb{P}}}}}\left(\mathcal{H} \right) \right)$ and ${\rm Re}_{U_{\mathcal{P}}}({\mathcal{H}})$ is the Rademacher complexity (Appendix-B from Supplemental Material (SM)). Let ${\mathcal{L}_{\widehat{d}}}({{\mathbb{{ P}}} }^{T,{\mathcal{X}}}_{t,j}, {{\mathbb{{ P}}}}^{S,{\mathcal{X}}}_{t,j})$ represent the Right-Hand Side (RHS) of Eq.~\eqref{colorllary_eq1}. We also assume that ${\mathcal{L}} \colon \mathcal{Y} \times \mathcal{Y} \to [0,1]$ is a symmetric and bounded loss function $\forall (y, y') \in {\mathcal{Y}^2},{\mathcal{L}} (y,y') \le {U'}$, and ${\mathcal{L}} ( \cdot, \cdot)$ obeys the triangle inequality, where ${ U'}$ is a positive number. 
\vspace{-10pt}

\vspace{-9pt}
\subsection{GB for a single model}
\label{GB_Single_Sec}
\vspace{-5pt}

Based on the definitions from Section~\ref{distributionShift}, we firstly derive the GB that can describe the learning process of a single model under TFCL.

\noindent \textBF{Theorem 1.}
\label{theorem1} Let $\mathcal{P}_i$ represent the distribution of all visited training samples (including all previous batches) drawn from $\mathcal{S}$ at $\mathcal{T}_i$. Let $h_{\mathcal{P}_i} = \arg \min_{h \in \mathcal{H}} {\mathcal{R}}( h,{\mathcal{P}}_i) $ and $h_{{\mathcal{M}}_i} = \arg \min_{h \in \mathcal{H}} {\mathcal{R}} (h, {\mathbb{P}}_{{\mathcal{M}}_i} )$ represent the ideal classifiers for ${\mathcal{P}}_i$ and ${\mathbb{P}}_{{\mathcal{M}}_i}$, respectively. We derive the GB between $\mathcal{P}_i$ and ${\mathbb{P}}_{{\mathcal{M}}_i}$, based on the results from Corollary 1~:
\begin{equation}
\begin{aligned}
{\mathcal{R}} \big( h,{\mathcal{P}}_i \big) &\le {\mathcal{R}} \big( h, h_{{\mathcal{M}}_i}, {\mathbb{P}}_{{\mathcal{M}_i}} \big) + {\mathcal{L}}_{\widehat{d}} \big( {\mathcal{P}}^{\mathcal{X}}_i,{\mathbb{P}}^{\mathcal{X}}_{{\mathcal{M}}_i} \big) + \eta \big( {\mathcal{P}}_i,{\mathbb{P}}_{{\mathcal{M}}_i} \big)  \,,
\label{theorem1_eq1}
\end{aligned}
\end{equation}
where $\eta \big( {\mathcal{P}}_i,{\mathbb{P}}_{{\mathcal{M}}_i} \big)$ is the optimal combined error ${\mathcal{R}}(h_{{\mathcal{P}}_i} ,h_{{\mathcal{M}}_i},{{\mathcal{P}}_i}) + {\mathcal{R}}(h_{{\mathcal{P}}_i} ,h_{{\mathcal{P}}_i}^\star,{{\mathcal{P}}_i})$ where ${\mathcal{R}}(  h_{{\mathcal{P}_i}},h_{{\mathcal{M}}_i} ,{\mathbb{P}}_{{\mathcal{M}}_i} )$ is the risk, expressed by ${\mathbb{E}}_{{{\bf x}} \sim {\mathbb{P}}_{{\mathcal{M}}_i}}[{\mathcal{L}} (h_{{\mathcal{P}}_i}({\bf x}),h_{{\mathcal{M}}_i}({\bf x}) ) ]$ and $h^\star_{{\mathcal{P}}_i}$ is the true labeling function for ${\mathcal{P}}_i$.

The proof is provided in Appendix-A from the SM. Compared to the GB used in the domain adaptation \cite{domainTheory}, Theorem 1 provides an explicit way to measure the gap between the model's prediction and the true labels in each training step (${\mathcal{T}}_i$). During the initial training stages (when $i$ is very small), the memory ${\mathcal{M}}_i$ can store all previous samples and GB is tight. However, as the number of training steps increases, the discrepancy distance $ {\mathcal{L}}_{\widehat{d}} \big( {\mathcal{{ P}}}^{\mathcal{X}}_i,{\mathbb{{ P}}}^{\mathcal{X}}_{{\mathcal{M}}_i} \big)$ would increase because ${\mathcal{M}}_i$ would lose the knowledge about previously learnt samples. This can lead to a degenerated performance on ${\mathcal{P}}_i$, corresponding to the forgetting process. Next we extend Theorem 1 to analyze the generalization performance on testing sets.

\noindent \textBF{Theorem 2.}
\label{theorem2}
For a given target domain ${\mathbb{P}}^T_{t,j}$, we derive the GB for a model at the training step ${\mathcal{T}}_i$~:
\begin{equation}
\begin{aligned}
{\mathcal{R}} \big( h, {\mathbb{P}}^T_{t,j}  \big) &\le {\mathcal{R}} \big( h, h_{{\mathcal{M}}_i}, {\mathbb{P}}_{{\mathcal{M}_i}} \big) + {\mathcal{L}}_{\widehat{d}} \big( {\mathbb{P}}^{T,{\mathcal{X}}}_{t,j},{\mathbb{P}}^{\mathcal{X}}_{{\mathcal{M}}_i} \big) + \eta \big( {\mathbb{P}}^T_{t,j},{\mathbb{P}}_{{\mathcal{M}}_i} \big)  \,,
\end{aligned}
\end{equation}
The proof is similar to that for Theorem 1. We can observe that the generalization performance on a target domain ${\mathbb{P}}^T_{t,j}$, by a model $h$ is relying mainly on the discrepancy distance between ${\mathbb{P}}^{T,{\mathcal{X}}}_{t,j}$ and ${\mathbb{P}}_{{\mathcal{M}}_i}$. In practice, we usually measure the generalization performance of $h$ on several data categories where each category is represented by a different underlying distribution. In the following, we extend Theorem 2 for multiple target distributions.

\noindent \textBF{Lemma 1.}
\label{lemma1}
For a given data stream $S =  \bigcup_{j=1}^{C^S_t} {\mathcal{D}^S_{t,j}}$ consisting of samples from ${\mathcal{D}}^S_t$, let ${\mathcal{D}}^T_t$ be the corresponding testing set and ${\mathbb{P}}^T_{t,j}$ represent the distribution of samples for the $j$-th category from ${\mathcal{D}}^T_{t}$, we derive the GB for multiple target domains as:
\vspace*{-0.1cm}
\begin{align}
\label{lemma1_eq1}
\sum\nolimits_{j = 1}^{C^T_{t}} \Big\{
{\mathcal{R}} \big( h, {\mathbb{P}}^T_{t,j}  \big) \Big\} &\le 
\sum\nolimits_{j = 1}^{C^T_t} \Big\{
{\mathcal{R}} \big( h, h_{{\mathcal{M}}_i}, {\mathbb{P}}_{{\mathcal{M}_i}} \big) + {\mathcal{L}}_{\widehat{d}} \big( {\mathbb{P}}^{T,{\mathcal{X}}}_{t,j},{\mathbb{P}}^{\mathcal{X}}_{{\mathcal{M}}_i} \big) + \eta \big( {\mathbb{P}}^T_{t,j},{\mathbb{P}}_{{\mathcal{M}}_i} \big) 
\Big\}
\,, 
\vspace*{-0.4cm}
\end{align}
where $C^T_t$ represents the number of testing data streams.

\textBF{Remarks.} Lemma 1 had the following observations~: \begin{inparaenum}[1)]
\item  The optimal performance of the model $h$ on the testing set can be achieved by minimizing the discrepancy distance between each target domain ${\mathbb{P}}^{T,{\mathcal{X}}}_{t,j}$ and the distribution ${\mathbb{P}}_{{\mathcal{M}}_i}$ at the training step ${\mathcal{T}}_i$.
\item \cite{OptimalCL} employs the set theory to theoretically demonstrate that a perfect memory is crucial for CL. In contrast, we evaluate the memory quality using the discrepancy distance in Eq.~\eqref{lemma1_eq1}, which provides a practical way to investigate the relationship between the memory and forgetting behaviour of existing approaches \cite{TinyLifelong,ContinualPrototype} at each training step without requiring any task information  (See more details in Appendix-D from SM).
\item \cite{LifelongInfinite} introduces a similar risk bound for forgetting analysis, which still requires the task information. In contrast, the proposed GB can be used in a more realistic CL scenario. Moreover, we provide the theoretical analysis for component diversity (Appendix-C from SM), which is missing in \cite{LifelongInfinite}.
\end{inparaenum} 

\vspace*{-7pt}
\subsection{GB for the dynamic expansion mechanism}

As discussed in Lemma 1, a memory of fixed capacity would lead to degenerated performance on all target domains. The other problem for a single memory system is the negative backward transfer \cite{GradientEpisodic} in which the performance of the model is decreased due to samples being drawn from entirely different distributions \cite{AchievingForgetting}. A Dynamic Expansion Model (DEM) can address these limitations from two aspects~:\begin{inparaenum}[1)]
\item DEM relieves the negative transfer by preserving the previously learnt knowledge into a frozen component from a mixture system;
\item DEM would achieve better generalization performance under TFCL by allowing each component to model one or only a few similar underlying data distributions.
\end{inparaenum} We derive GB for DEM and show the advantages of DEM for TFCL.

\noindent \textBF{Definition 7.}
\textBF{(Dynamic expansion mechanism.)} Let ${\mathcal{G}}$ represent a dynamic expansion model and ${G}_j$ represent the $j$-th component in ${\mathcal{G}}$, implemented by a single classifier. ${\mathcal{G}}$ starts with training its first component during the initial training phase and would add new components during the following training steps. In order to overcome forgetting, only the newly created component is updated each time, while all previously trained components have their parameters frozen. 

\noindent \textBF{Theorem 3.} For a given data stream ${\mathcal{S}} = \{{\bf X}^b_1,\cdots, {\bf X}^b_n \} $, let ${\mathcal{P}}_{(i,j)}$ represent the distribution of the $j$-th training batch ${\bf X}^b_j$ (visited) drawn from ${\mathcal{S}}$ at $\mathcal{T}_i$. We assume that ${\mathcal{G}} = \{ G_1,\cdots,G_c \}$ trained $c$ components at $\mathcal{T}_i$. Let ${\mathcal{T}} = \{ {\mathcal{T}}_{k_1},\cdots,{\mathcal{T}}_{k_c} \}$ be a set of training steps, where $G_j$ was frozen at $\mathcal{T}_{k_j}$. We derive the GB for $\mathcal{G}$ at ${\mathcal{T}}_i$ as:
\begin{equation}
\begin{aligned}
\frac{1}{i}\sum\nolimits_{j = 1}^i \big\{ {\mathcal{R}} \big( h,{\mathcal{P}}_{(i,j)} \big)  \big\} 
 &\le 
 \frac{1}{i}\sum\nolimits_{j = 1}^i \big\{
 {\rm F}_{S} \big( {\mathcal{P}}_{(i,j)} , {\mathcal{G}} \big) \big\} \,,
\label{theorem3_eq1}
\end{aligned}
\end{equation}
\noindent where ${\rm F}_{S}(\cdot,\cdot)$ is the selection function, defined as~:
\begin{align}
\label{theorem3_eq2}
{\rm F}_{S} \big( {\mathcal{P}}_{(i,j)} , {\mathcal{G}} \big) &\buildrel \Delta \over = 
\mathop { \min }\limits_{ k_1,\cdots,k_c } \Big\{
{\mathcal{R}} \big( h, h_{{\mathcal{M}}_{k_i}}, {\mathbb{P}}_{{\mathcal{M}_{k_i}}} \big) 
+ {\mathcal{L}}_{\widehat{d}} \big( {\mathcal{P}}^{\mathcal{X}}_{(i,j)},{\mathbb{P}}^{\mathcal{X}}_{{\mathcal{M}}_{k_i}} \big)+ \eta \big( {\mathcal{P}}_{(i,j)},{\mathbb{P}}_{{\mathcal{M}}_{k_i}} \big)
\Big\} 
\,,
\end{align}
where ${\mathbb{P}}_{{\mathcal{M}_{k_i}}}$ represents the memory distribution at $\mathcal{T}_{k_i}$.\vspace{-1.0pt} The proof is provided in Appendix-B from SM. It notes that ${\rm F}_{S}(\cdot,\cdot)$ can be used for arbitrary distributions. We assume an ideal model selection in Eq.~\eqref{theorem3_eq2}, where always the component with the minimal risk is chosen. DEM can achieve the minimal upper bound to the risk (Left Hand Side (LHS) of Eq.~\eqref{theorem3_eq1}) when comparing with a single model (Theorem 3),. Then, we derive the GB for analyzing the generalization performance of ${\mathcal{G}}$ on multiple target distributions.

\noindent \textBF{Lemma 2.}
\label{lemma2}
For a given data stream $S =  \bigcup ^{C^S_t}_{j} {\mathcal{D}^S_{t,j}}$ consisting of samples from ${\mathcal{D}}^S_{t}$, we have a set of target sets $\{ {\mathcal{D}}^T_{t,1}, \cdots,{\mathcal{D}}^T_{t,C^T_t}  \}$, where each ${\mathcal{D}}^T_{t,j}$ contains $C^{b}_{t,j}$ batches of samples.\vspace{-2.0pt} Let ${\mathbb{P}}^T_{t,j}(d)$ represent the distribution of the $d$-th batch of samples in ${\mathcal{D}}^T_{t,j}$. We assume that $\mathcal{G}$ consists of $c$ components trained on samples from $\mathcal{S}$, at ${\mathcal{T}}_i$. We derive the GB for multiple target domains as~:
\begin{align}
\sum\nolimits_{j = 1}^{C^T_{t}} \Big\{
\sum\nolimits_{d = 1}^{C^b_{t,j}} 
{\mathcal{R}} \big( h, {\mathbb{P}}^T_{t,j}(d)  \big) \Big\} &\le 
\sum\nolimits_{j = 1}^{C^T_{t}} \Big\{
\sum\nolimits_{d = 1}^{C^b_{t,j}} 
\big\{ {\rm F}_{S} \big( {\mathbb{P}}^T_{t,j}(d) , {\mathcal{G}} \big) \big\}
\Big\}.
\label{Lemma2_eq1}
\end{align}
\noindent \textBF{Remark.} We have several observations from Lemma 2~:
\begin{inparaenum}[1)]
\item The generalization performance of $\mathcal{G}$ is relying on the discrepancy distance between each target distribution ${\mathbb{P}}^T_{t,j}$ and the memory distribution ${\mathbb{P}}_{{\mathcal{M}}_{k_i}}$ of the selected component (Also see details in Appendix-C from SM).
\item Eq.~\eqref{Lemma2_eq1} provides the analysis of the trade-off between the model's complexity and generalization for DEM \cite{NeuralDirichelt,LifelongUnsupervisedVAE}. By adding new components, $\mathcal{G}$ would capture additional information of each target distribution and thus improve its performance. On the other hand, the selection process ensures a probabilistic diversity of stored information, aiming to capture more knowledge with a minimal number of components. 
\end{inparaenum} 

In practice, we usually perform the model selection for ${\mathcal{G}}$ by using a\vspace{-1.0pt} certain criterion that only accesses the testing samples without task labels. Therefore, we introduce a selection criterion ${\widehat{\rm F}}(\cdot,\cdot)$, implemented by comparing the sample log-likelihood~:
\begin{align}
{\widehat{\rm F}} \big( {\mathbb{P}}^T_{t,j}(d) , {\mathcal{G}} \big) &\buildrel \Delta \over =
\mathop {\arg \max }\limits_{ k_1,\cdots,k_c } 
\big\{
{\mathbb{E}}_{{\bf x} \sim {\mathbb{P}}^T_{t,j}(d) }
[ \hat{f} ( {\bf x}, G_{k_j} )]
\big\},
\label{model_selection_eq1}
\end{align}
where $j=\{1,\cdots,c \}$ and ${\hat f}(\cdot,\cdot)$ is a pre-defined sample-log likelihood function. Then Eq.~\eqref{model_selection_eq1} is used for model selection:
\begin{align}
\label{lemma2_5}
{\widehat{\rm F}}_{S} \big( {\mathbb{P}}_{t,j}^{T}(d) , {\mathcal{G}} \big) &= \big\{
{\mathcal{R}} \big( h, h_{{\mathcal{M}}_{s}}, {\mathbb{P}}_{{\mathcal{M}_{s}}} \big) 
\notag +  {\mathcal{L}}_{\widehat{d}} \big( {\mathbb{P}}^{T,{\mathcal{X}}}_{t,j}(d),{\mathbb{P}}^{\mathcal{X}}_{{\mathcal{M}}_{s}} \big)+ \eta \big( {\mathbb{P}}^T_{t,j}(d),{\mathbb{P}}_{{\mathcal{M}}_{s}} \big)
\,|\, \notag \\&s = {\widehat{ \rm F}} \big( {\mathbb{P}}^T_{t,j}(d), {\mathcal{G}} \big) \big\}
\,.
\end{align}
We rewrite Eq.~\eqref{Lemma2_eq1} using Eq.~\eqref{lemma2_5}, resulting in~:
\begin{align}
\sum\nolimits_{j = 1}^{C^T_{t}} \Big\{&
\sum\nolimits_{d = 1}^{C^b_{t,j}} 
{\mathcal{R}} \big( h, {\mathbb{P}}^T_{t,j}(d)  \big) \Big\} \le 
\sum\nolimits_{j = 1}^{C^T_{t}} \Big\{
\sum\nolimits_{d = 1}^{C^b_{t,j}} 
\big\{
 {\widehat{\rm F}}_{S} \big( {\mathbb{P}}_{t,j}^{T}(d) , {\mathcal{G}} \big) 
\big\}
\Big\}
\,.
\label{lemma2_eq2}
\end{align}
Compared with an ideal solution (Eq.~\eqref{Lemma2_eq1}), Eq.~\eqref{lemma2_eq2} would involve the extra error terms caused by the selection process (Eq.~\eqref{lemma2_5}), expressed as $\sum\nolimits_{j = 1}^{C^T_{t}} \Big\{
\sum\nolimits_{d = 1}^{C^b_{t,j}}
\big\{
 {{\widehat{\rm F}}}_{S} \big( {\mathbb{P}}_{t,j}^T(d) , {\mathcal{G}} \big)  -
 { {\rm F}}_{S} \big( {\mathbb{P}}_{t,j}^T(d), {\mathcal{G}} \big) 
\big\}
\Big\}$. In Section~\ref{sec:Methodology}, we introduce a new CL framework according to the theoretical analysis. 

\vspace{-10pt}
\subsection{Parallels with other studies defining lifelong learning bounds}
\vspace{-5pt}
In this section, we discuss the differences between the results in this paper and those from other studies proposing lifelong learning bounds. The bound (Theorem 1) in \cite{Theory_drifiting} assumes that the model is trained on all previous samples, which is not practical under a TFCL setting. In contrast, in Theorem 1 from this paper, the model is trained on a memory buffer, which can be used in the context of TFCL. In addition, the bounds in \cite{Theory_drifiting} mainly provide the theoretical guarantees for the performance when learning a new data distribution (Theorem 1 from \cite{Theory_drifiting}), which does not provide an analysis of the model's forgetting. In contrast, this paper studies the forgetting analysis of our model and its theoretical developments can be easily extended for analyzing the forgetting behaviour of a variety of continual learning methods, while this cannot be said about the study from \cite{Theory_drifiting} (See Appendix-D from supplemental material).

When comparing with \cite{LifelongIID}, our theoretical analysis does not rely on explicit task boundaries, which is a more realistic assumption for TFCL. In addition, \cite{LifelongIID} employs the KL divergence to measure the distance between two distributions, which would require knowing the explicit distribution form and is thus hard to evaluate in practice during the learning. However, our theory employs the discrepancy distance, which can be reliably estimated. Moreover, the study from \cite{LifelongIID} does not develop a practical algorithm to be used according to the theoretical analysis. In contrast, our theory provides guidelines for the algorithm design under the TFCL assumption. Finally, inspired by the proposed theoretical analysis, this paper develops a successful continual learning approach for TFCL.

Theorem 1 in \cite{minMaxLearning}, similarly to \cite{Theory_drifiting}, assumes that the model is trained on uncertain data sets over time, which would not be suitable for TFCL since the model can not access previously learnt samples under TFCL. In contrast, Theorem 1 in our theory represents a realistic TFCL scenario in which the model is trained on a fixed-length memory buffer and is evaluated on all previously seen samples. Therefore,  the analysis for the forgetting behaviour of the model under TFCL relies on Theorem 1 and its consequences. In addition,  the study from \cite{minMaxLearning}, similarly to \cite{Theory_drifiting,LifelongIID},  only provides a theoretical guarantee for a single model. In contrast, we extend our theoretical analysis to the dynamic expansion model, which motivates us to develop a novel continual learning approach for TFCL. Moreover, this paper is the first work to provide the theoretical analysis for the diversity of knowledge recorded by different components (See Appendix-C from supplemental material). This analysis indicates that by maintaining the knowledge diversity among different components we ensure a good trade-off between the model's complexity and generalization performance, thus providing invaluable insights into algorithm design for TFCL.

\section{Methodology}
\label{sec:Methodology}

\subsection{Network architecture}

Each component $G_j \in {\mathcal{G}}$ consists of a classifier $h_j$ implemented by a neural network $f_{\varsigma_j}({\bf x})$ with trainable parameters $\varsigma_j$, and a variational autoencoder (VAE) model implemented by an encoding network ${\rm enc}_{\omega _j}$ as well as the corresponding decoding network ${\rm dec}_{\theta_j}$, with trainable parameters $\{\omega_j,\theta_j \}$. Due to its robust generation and inference mechanisms, VAE has been widely used in many applications \cite{MixtureOfVAEs,DeepMixtureVAE,InfoVAEGAN_conference,JontLatentVAEs,LifelongInfinite,InfoVAE}. This paper employs a VAE for discrepancy estimation and component selection. The loss function for the $j$-th component at $\mathcal{T}_i$ is defined as~:
\begin{equation}
\begin{aligned}
{\mathcal{L}}_{class}(G_j,{\mathcal{M} }_i)  \buildrel \Delta \over =  \frac{1}{{| {\mathcal{M}}_i |}}\sum\nolimits_{t = 1}^{|{\mathcal{M}}_i|} \big\{ {\mathcal{L}}_{ce} \big( h_j( {\bf x}_t), { y}_t \big) \big\}  \,,
\label{classifierLoss_eq}
\end{aligned}
\end{equation}
\vspace{-5pt}
\begin{equation}
\begin{aligned}
{\mathcal{L}}_{VAE}(G_j,{\mathcal{M} }_i)  \buildrel \Delta \over =  \; & {\mathbb{E}_{{q_{{\omega _j}}}({\bf{z}} \,|\, {\bf{x}})}}\left[ {\log {p_{{\theta _j}}}({\bf{x}}_t \,|\, {\bf{z}})} \right] - {D_{KL}}\left[ {{q_{{\omega _j}}}({\bf{z}} \,|\, {\bf{x}}_t) \,||\, p({\bf{z}})} \right],
\label{VAEloss_eq}
\end{aligned}
\end{equation}
where $\{{\bf x}_t,y_t \}\sim \mathcal{M}_i$, where $|\mathcal{M}_i|$ is the memory buffer size, and ${\mathcal{L}}_{ce}(\cdot)$ is the cross-entropy loss. $\mathcal{L}_{VAE}(\cdot,\cdot)$ is the VAE loss \cite{VAE}, ${p_{{\theta _j}}}({\bf{x}}_t \,|\, {\bf{z}})$ and ${q_{{\omega _j}}}({\bf{z}} \,|\, {\bf{x}}_t)$ are the encoding and decoding distributions, implemented by ${\rm enc}_{\omega _j}$ and ${\rm dec}_{\theta_j}$, respectively.  We also implement $\hat{f}(\cdot,\cdot)$ in Eq.~\eqref{model_selection_eq1} by $-\mathcal{L}_{VAE}(\cdot,\cdot)$ for the component selection at the testing phase. The training algorithm for the proposed ODDL consists of three stages (See more information in Appendix-E from SM). In the initial training stage, we aim to learn the initial knowledge of the data stream and preserve it into a frozen component ${G}_1$, which can provide information for the dynamic expansion and sample selection evaluation in the subsequent learning. During the evaluator training stage we train the current component as the evaluator that estimates the discrepancy distance between each previously learnt component and the memory buffer, providing appropriate signals for the model expansion. In the sample selection stage, we train the current component with new data, while we aim to promote knowledge diversity among components.

In the following, we firstly propose a new dynamic expansion mechanism based on the discrepancy criterion. Then we introduce a new sample selection approach that can further improve the performance of the model. Finally, we provide the detailed algorithm implementation.

\subsection{Discrepancy based mixture model expansion}
\label{section_dynamicExpansi}

From Lemma 2, we observe that the probabilistic diversity of trained components in $\mathcal{G}$ can ensure a compact network architecture while maintaining a good generalization performance (See Theorem 4 in Appendix-C from SM). In order to achieve this, we maximize the discrepancy between each trained component of ${\mathcal{G}}$ and the current memory buffer, during the training, by using the discrepancy distance (notations are defined in \textBF{Theorem 3}), expressed as~:
\begin{equation}
\begin{aligned}
{\mathcal{M}}^\star = 
\mathop {\arg \max }\limits_{{\mathcal{M}}_i} 
\sum\nolimits_j^{\rm{c}} \big\{
{\mathcal{L}}_{\widehat{d}} ( {\mathbb{P}}^{\mathcal{X}}_{{\mathcal{M}}_{k_j}}, {\mathbb{P}}^{\mathcal{X}}_{{\mathcal{M}}_i}  ) \big\}\,,
\label{optimization_eq}
\end{aligned}
\end{equation}
where $i=k_{c} + 1,\cdots,n$ represents the index of the training steps and $k_c$ is the $c$-th component trained at ${\mathcal{T}}_{k_c}$. ${\mathcal{M}}^\star$ is the optimal memory. Eq.~\eqref{optimization_eq} can be seen as a recursive optimization problem when ${\mathcal{G}}$ dynamically adds new components ($c$ is increased) during the training. In order to provide a practical way to solve Eq.~\eqref{optimization_eq} while balancing the model's complexity and performance, we derive an expansion criterion based on the discrepancy distance~:
\begin{equation}
\begin{aligned}
\min \Big\{
{\mathcal{L}}_{\widehat{d}} \big({\mathbb{P}}^{\mathcal{X}}_{{\mathcal{M}}_{k_j}}, {\mathbb{P}}^{\mathcal{X}}_{{\mathcal{M}}_i} \big) \,|\,  j = 1,\cdots,c \Big\} \ge \lambda \, ,
\label{multi_expansion_eq2}
\end{aligned}
\end{equation}
where $\lambda \in [0,4]$ is an architecture expansion threshold. If the current memory distribution ${\mathbb{P}}_{{\mathcal{M}}_i}$ is sufficiently different from each component (satisfies Eq.~\eqref{multi_expansion_eq2}), ${\mathcal{G}}$ will add a new component to preserve the knowledge of the current memory ${\mathcal{M}}_i$, while also encouraging the probabilistic diversity among the trained components.

In the following, we describe the implementation. We start by training the first component $G_1$ which consists of a classifier $h_1$ (classification task) and a VAE model $v_1$ (model selection at the testing phase). We also train an additional component called the evaluator $G_e = \{h_e,v_e\}$, at the initial training stage, which aims to capture the future information about the data stream. Once the memory $\mathcal{M}_j$ reaches its maximum size at $\mathcal{T}_j$, we freeze the weights of the first component to preserve the previously learnt knowledge while continually training the evaluator during the following training steps. Since the evaluator continually captures the knowledge from the current memory ${\mathcal{M}}_i$, we check the model expansion using Eq.~\eqref{multi_expansion_eq2} at ${\mathcal{T}}_i$~:
\begin{equation}
\begin{aligned}
{\mathcal{L}}^\star_d \big({\mathbb{P} }^{{\mathcal{X}}}_{{\mathcal{M}}_{k_1}}, {\mathbb{P}}^{\mathcal{X}}_{{\mathcal{M}}_i} \big) \ge \lambda \, .
\label{expansion_eq1}
\end{aligned}
\end{equation}
It notes that we can not access the previously learnt memory distribution\vspace{-1pt} ${\mathbb{P}}^{\mathcal{X}}_{{\mathcal{M}}_{k_1}}$ and we approximate it by\vspace{-1.5pt} using the auxiliary distribution ${\mathbb{P} }^{{\mathcal{X}}}_{v^j_1}$ formed by\vspace{-1pt} samples drawn from $v_1^j$ of $G_1^j$, where the superscript $j$ represents the first component finishing the training at ${\mathcal{T}}_j$. ${\mathcal{L}}^\star_d (\cdot, \cdot )$ is the estimator of the discrepancy distance, achieved by $h^i_e$ and $h^j_1$.
\begin{equation}
\begin{aligned}
{\mathcal{L}}^\star_d \big({\mathbb{P} }^{{\mathcal{X}}}_{v^j_1}, {\mathbb{P}}_{{\mathcal{M}}_i} \big) &\buildrel \Delta \over = \Big| {\mathbb{E}}_{{\bf x} \sim {\mathbb{P} }^{{\mathcal{X}}}_{v^j_1} }  \big[ {\mathcal{L}} \big( h^j_1 ({\bf x}),h^i_{e} ({\bf x}) \big)   \big] - 
{\mathbb{E}}_{{\bf x} \sim {\mathbb{P}}_{{\mathcal{M}}_i} }  \big[ {\mathcal{L}} \big( h^j_1 ({\bf x}),h^i_{e} ({\bf x}) \big)   \big]
\Big|.
\label{discrepancyForTwo}
\end{aligned}
\end{equation}

\begin{wrapfigure}{r}{7cm}
\centering
\vspace{-13pt}
\includegraphics[width=0.5\textwidth]{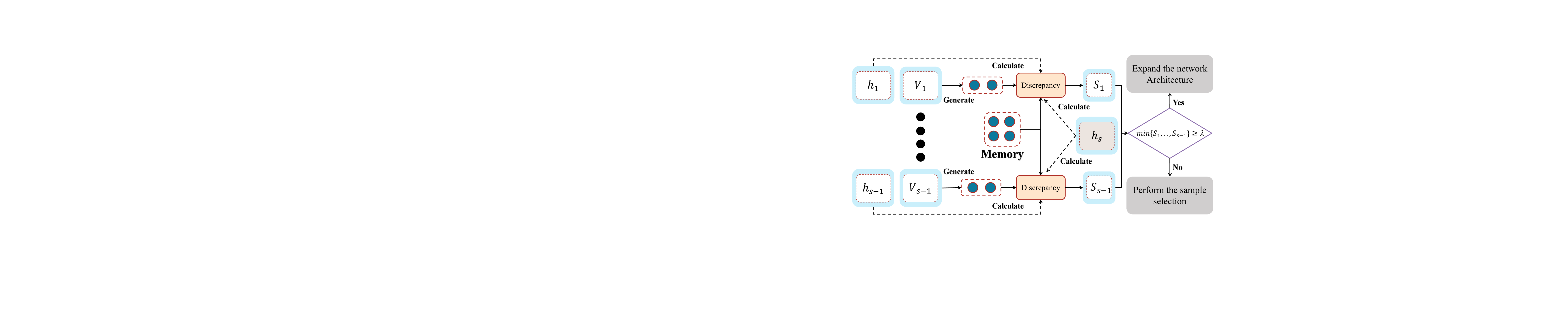}
\caption{We generate the knowledge by using the VAE ($V_j$) of each previous component $G_j, j =1, \cdots,s-1$, which is used to evaluate the discrepancy distance $S_j = {\mathcal{L}}^\star_d \big({\mathbb{P} }^{{\mathcal{X}}}_{v_j}, {\mathbb{P}}_{{\mathcal{M}}_i} \big)$ at ${\mathcal{T}}_i$ (Eq.~\eqref{discrepancyForTwo}) between $G_j$ and the memory buffer. Then we use these discrepancy scores $\{S_1,\cdots,S_{s-1}\}$ to check the model expansion (Eq.~\eqref{multi_expansion_eq1})}
\label{small_structure}
\end{wrapfigure}

If Eq.~\eqref{expansion_eq1} is fulfilled, then we add $G^i_{e}$ into the model $\mathcal{G}$ while building a new evaluator $(G^{i+1}_{e} = G_3^{i+1})$ at $\mathcal{T}_{i+1}$; otherwise, we train $G^i_{e} \to G^{i+1}_{e}$ at the next training step, $\mathcal{T}_{i+1}$. Furthermore, for satisfying the diversity of knowledge in the components from the mixture model (See Lemma 2), we clear the memory $\mathcal{M}_i$ when performing the expansion in order to ensure that we would learn non-overlapping distributions during the following training steps. We can also extend Eq.~\eqref{expansion_eq1} to the expansion criterion for $\mathcal{G} = \{G_1,\cdots,G_{s-1}\}$ that has already preserved $(s-1)$ components, illustrated in Fig.~\ref{small_structure}:
\begin{equation}
\begin{aligned}
\min \Big\{
{\mathcal{L}}^\star_d \big({\mathbb{P} }^{{\mathcal{X}}}_{v^{k_j}_j}, {\mathbb{P}}^{\mathcal{X}}_{{\mathcal{M}}_i} \big) \,|\,  j = 1,\cdots,s-1 \Big\} \ge \lambda \, ,
\label{multi_expansion_eq1}
\vspace{-20pt}
\end{aligned}
\end{equation}
where ${\mathbb{P} }^{{\mathcal{X}}}_{v^{k_j}_j}$ is the distribution of the generated samples\vspace{-3pt} by $G^{k_j}_j$ and denote $S_j= {\mathcal{L}}^\star_d \big({\mathbb{P} }^{{\mathcal{X}}}_{v^{k_j}_{j}}, {\mathbb{P}}_{{\mathcal{M}}_i} \big)$.

\subsection{Sample selection}

According to Lemma 2, the probabilistic diversity of the knowledge accumulated in the components is crucial for the performance. In the following, we also introduce a novel sample selection approach from the memory buffer that further encourages component diversity. The primary motivation behind the proposed sample selection approach is that we desire to store in the memory buffer those samples that are completely different from the data used for training the other components. This mechanism enables the currently created component to capture a different underlying data distribution. 
To implement this goal, we estimate the discrepancy distance on a pair of samples as the diversity score~:
\begin{equation}
\begin{aligned}
{\cal L}_d^s({\cal G},{{\bf{x}}_c}){\rm{ }} &\buildrel \Delta \over = \frac{1}{s}\sum\nolimits_{t = 1}^s \big|{\cal L}(h_t^{k_t}({\bf{x}}_c),h_{e}^i({\bf{x}}_c)) - {\cal L}(h_t^{k_t}({{\bf{x}}^{k_t}_t}),h_{e}^i({\bf{x}}^{k_t}_t)) \big| \,,
\label{diversityScore}
\end{aligned}
\end{equation}
where ${\bf x}_c$ and ${\bf{x}}^{k_t}_t$ are the $c$-th sample from $\mathcal{M}_i$ and the generated one drawn from $G^{k_t}_t$, respectively. $h_t^{k_t}({\bf{\cdot }})$ is the classifier of $G^{k_t}_t$ in $\mathcal{G}$. Eq.~\eqref{diversityScore} evaluates the average discrepancy distance between the knowledge generated by each trained component and the stored samples, which guides for the sample selection at the training step (${\mathcal{T}}_i$) as~:
\begin{equation}
\begin{aligned}
\vspace{-10pt}
{\mathcal{M}}_i =  \bigcup\nolimits^{|{{ \mathcal{M}}}'_i| - b}_{j=1} {{\mathcal{M}}}'_i[j]\,,
\label{sampleSelection}
\end{aligned}
\end{equation}
where ${{\mathcal{M}}}'_i$ is the sorted memory that satisfies the condition ${\mathcal{L}}^s_d({\mathcal{G}}, {{\mathcal{M}}}'_i[a] ) > {\mathcal{L}}^s_d({\mathcal{G}}, {{\mathcal{M}}}'_i[q] )$ for $a < q$. ${{\mathcal{M}}}'_i[j]$ represents the $j$-th stored sample and $b=10$ is the batch size. We name the approach with sample selection as ODDL-S.

\begin{figure*}[t]
\hspace{-5pt}
	\centering
	\subfigure[Risk on the training set.]{
		\centering
		\includegraphics[scale=0.325]{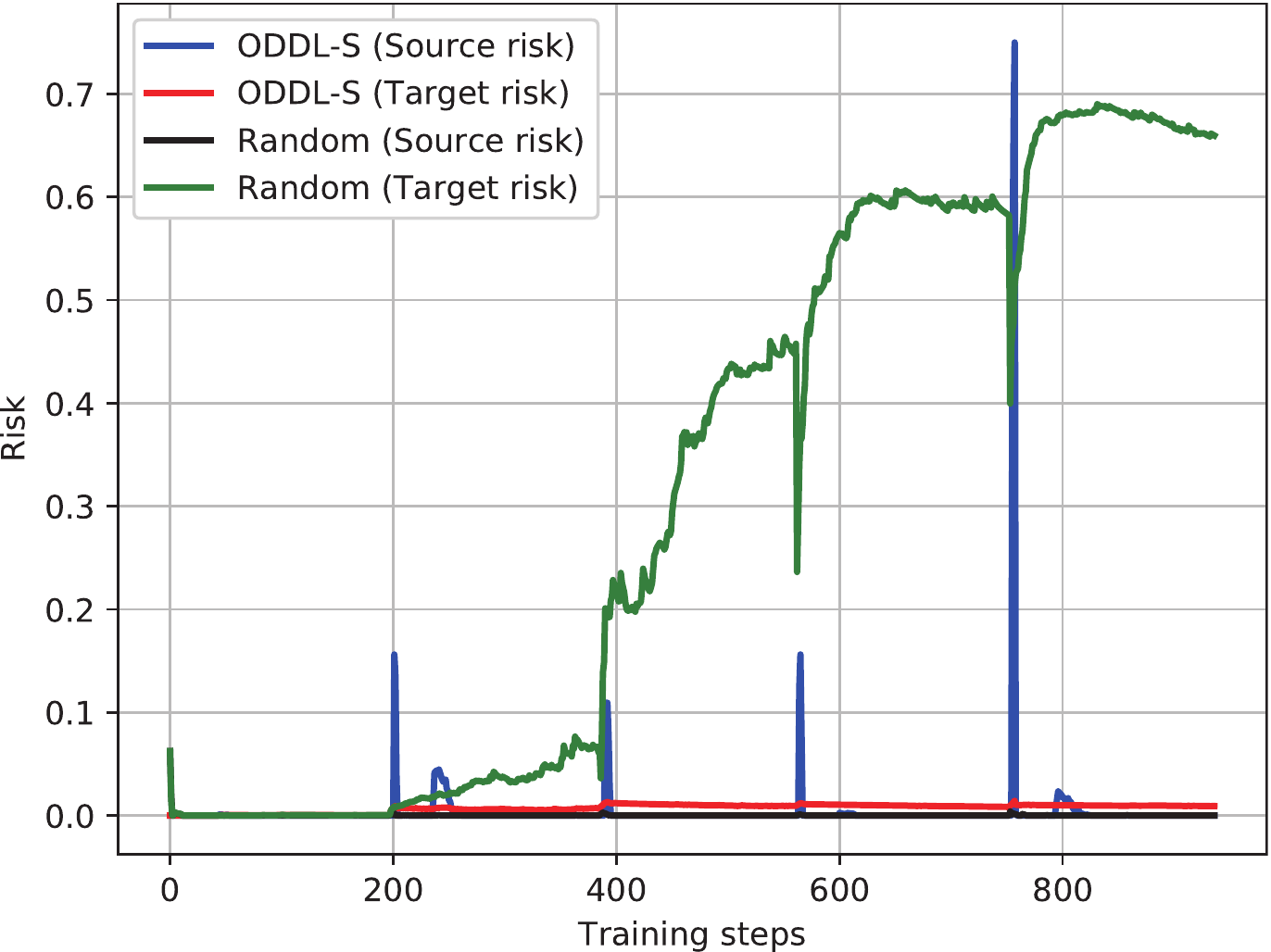}
	}
\subfigure[Risk on all target sets.]{
		\centering
		\includegraphics[scale=0.325]{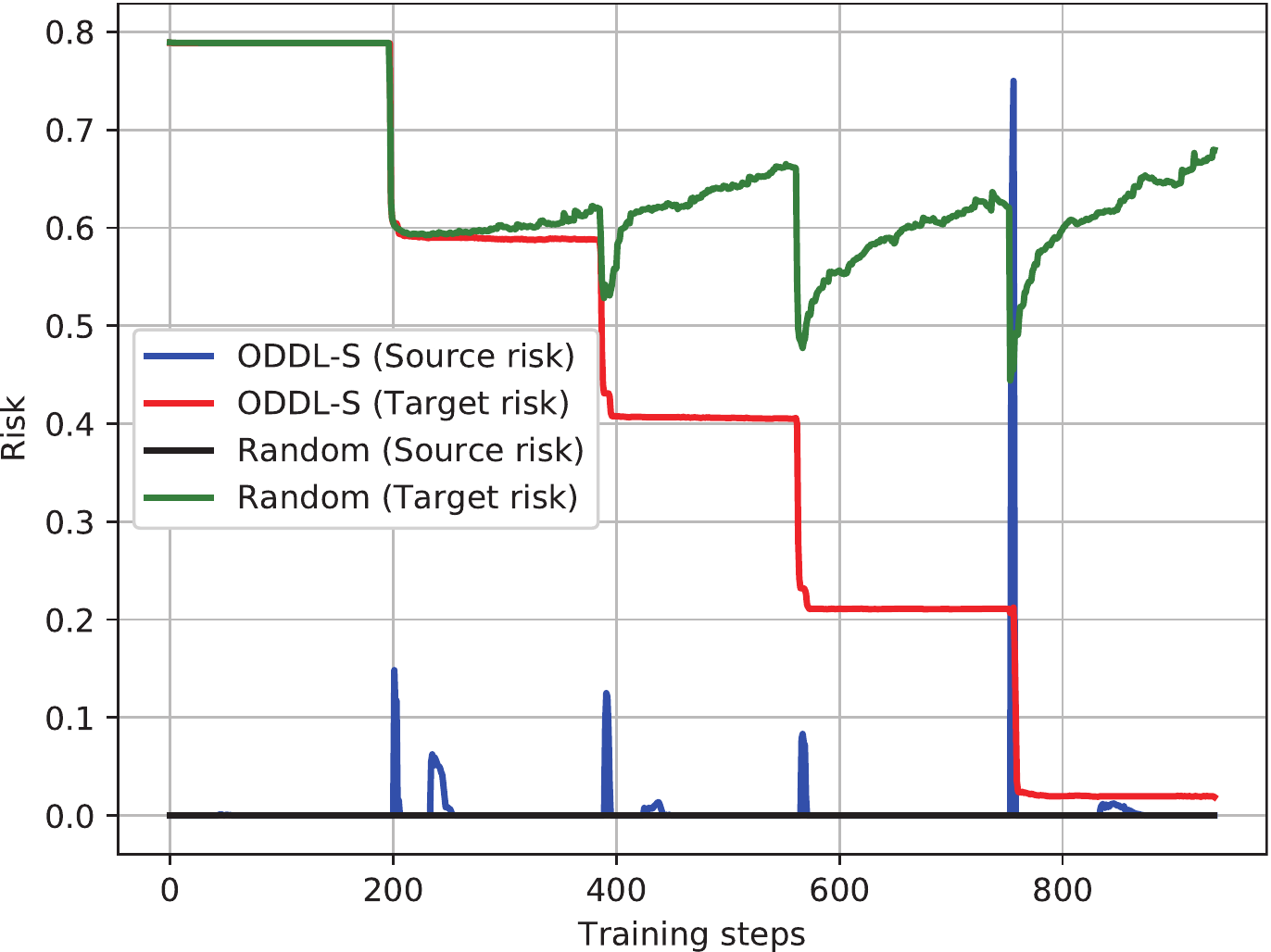}
	}
	\subfigure[Results on MNIST.]{
		\centering
		\includegraphics[scale=0.325]{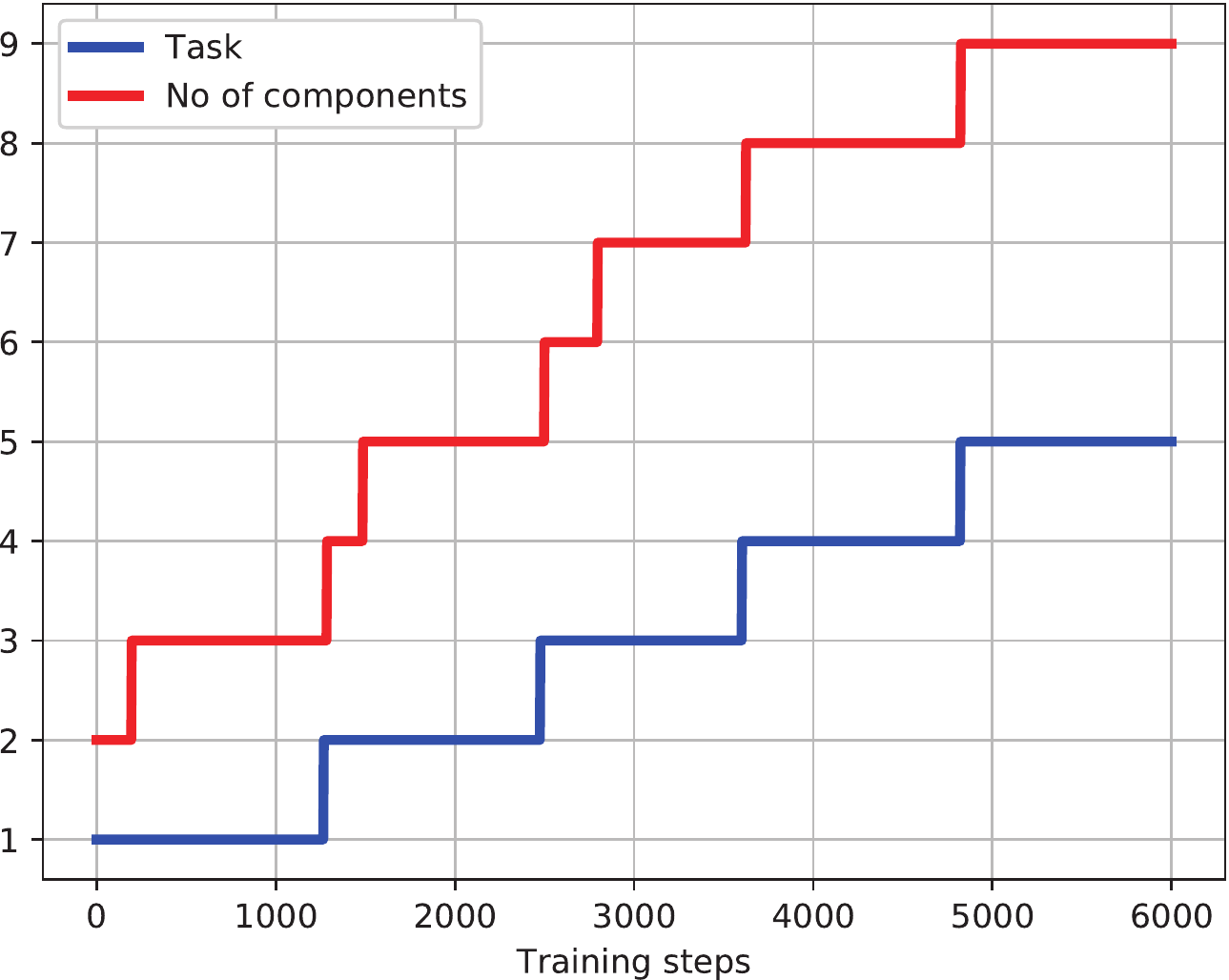}
	}
	\vspace{-5pt}
	\caption{The forgetting analysis of a single model and DEM in (a) and (b) where the batch size is 64 in the data stream. Data distribution shift and increasing the number of components in ODDL-S during the training in (c).}
	\label{riskResult2}
	\vspace{-15pt}
\end{figure*}

\subsection{Algorithm implementation}

We provide the pseudocode (\textBF{Algorithm 1}) in Appendix-E from SM. The algorithm has three main stages~:
\begin{itemize}[leftmargin=10pt]
\setlength{\itemsep}{1pt}
\setlength{\parsep}{1pt}
\setlength{\parskip}{1pt}

\item \textBF{(Initial training).} We start by building two components $\{G_1,G_2 \}$ where we only add $G_1$ in $\mathcal{G}$ and consider $G_2  = G_{e} $ as the Evaluator.
$G_1$ and $G_2$ are trained jointly using Eq.~\eqref{VAEloss_eq} and Eq.~\eqref{classifierLoss_eq} during the initial training stage until the memory $\mathcal{M}_j$ reaches its maximum size $|{\mathcal{M}}|^{max}$ at a certain training step ${\mathcal{T}}_j$. Then we freeze the first component $G_1$ and perform the second stage.
\item \textBF{(Evaluator training).} In this stage, we only update the Evaluator using Eq.~\eqref{VAEloss_eq} and Eq.~\eqref{classifierLoss_eq} on ${\mathcal{M}}_{j+1}$ at ${\mathcal{T}}_{j+1}$. If $ |\mathcal{M}_{j+1}| \ge 
|\mathcal{M}|^{max}$, then we evaluate the discrepancy distance using Eq.~\eqref{discrepancyForTwo} to check the expansion (Eq.~\eqref{multi_expansion_eq1}). If the expansion criterion is satisfied then we add $G_{e}$ to $\mathcal{G}$ and build a new Evaluator while cleaning up the memory $\mathcal{M}_{j+1}$, otherwise, we perform the sample selection.
\item \textBF{(Sample selection).} We evaluate the diversity score for each stored sample in $\mathcal{M}_{j+1}$ using Eq.~\eqref{diversityScore}. We then perform the sample selection for $\mathcal{M}_{j+1}$ using Eq.~\eqref{sampleSelection} and return back to the second stage.
\end{itemize} 

\section{Experiments}

We perform the experiments to address the following research questions: \begin{inparaenum}[1)]
\item What factors would cause the model's forgetting, and how to explain such behaviour? 
\item How efficient is the proposed ODDL-S under TFCL benchmarks? 
\item How important is each module in OODL-S?
\end{inparaenum}
\vspace*{-0.1cm}

In this experiment, we adapt the TFCL setting from \cite{ContinualPrototype} which employs several datasets including Split MNIST \cite{MNIST}, Split CIFAR10 \cite{CIFAR10} and Split CIFAR100 \cite{CIFAR10}. The detailed information for datasets, hyperparameters and network architectures is provided in Appendix-F from the supplementary material. The code is available at \url{https://dtuzi123.github.io/ODDL/}.

\begin{wraptable}{r}{6.7cm}
\vspace{-16pt}
\centering
	\caption{The accuracy of various continual learning models for five independent runs.}
	\vspace{5pt}
\renewcommand\arraystretch{1.15}
\scriptsize
\setlength{\tabcolsep}{0.97mm}{\begin{tabular}{@{}l cc cc c@{} } 
\toprule \textBF{Methods}   &\textBF{Split MNIST}& \textBF{Split CIFAR10} &\textBF{Split CIFAR100}  \\
\midrule 
	  finetune*&19.75 $\pm$ 0.05&18.55 $\pm$ 0.34&3.53 $\pm$ 0.04 \\
	  GEM* &93.25 $\pm$ 0.36&24.13 $\pm$ 2.46&11.12 $\pm$ 2.48 \\
	  iCARL*  &83.95 $\pm$ 0.21&37.32 $\pm$ 2.66&10.80 $\pm$ 0.37 \\
	  reservoir*  &92.16 $\pm$ 0.75&42.48 $\pm$ 3.04&19.57 $\pm$ 1.79 \\
	 MIR*  &93.20 $\pm$ 0.36&42.80 $\pm$ 2.22&20.00 $\pm$ 0.57 \\
	 GSS*  &92.47 $\pm$ 0.92&38.45 $\pm$ 1.41&13.10 $\pm$ 0.94 \\
	CoPE-CE* &91.77 $\pm$ 0.87&39.73 $\pm$ 2.26&18.33 $\pm$ 1.52 \\
	CoPE*  &93.94 $\pm$ 0.20&48.92 $\pm$ 1.32&21.62 $\pm$ 0.69 \\
		ER + GMED$\dag$&82.67 $\pm$ 1.90& 34.84 $\pm$ 2.20&20.93 $\pm$ 1.60   \\
	  ER$_{a}$ + GMED$\dag$ &82.21 $\pm$ 2.90&47.47 $\pm$ 3.20&19.60 $\pm$ 1.50  \\
	CURL*  &92.59 $\pm$ 0.66&-&- \\
	  CNDPM*  &93.23 $\pm$ 0.09&45.21 $\pm$ 0.18&20.10 $\pm$ 0.12 \\
	  	  Dynamic-OCM& 94.02 $\pm$ 0.23 & 49.16  $\pm$ 1.52 & 21.79 $\pm$ 0.68\\
	 \hline
	 \hline
	   ODDL& 94.85  $\pm$ 0.02 & 51.48 $\pm$ 0.12 & 26.20 $\pm$ 0.72  \\
	   ODDL-S & \textBF{95.75}  $\pm$ 0.05 &\textBF{52.69} $\pm$ 0.11 & \textBF{27.21} $\pm$ 0.87 \\
\bottomrule 
\end{tabular}
\label{classification}
}
\vspace{-5pt}
\end{wraptable}

\subsection{Empirical results for the forgetting analysis}

In this section, we investigate the forgetting behaviour of the model according to the proposed theoretical framework. Firstly, we train a single classifier $h$ under Split MNIST database, as a baseline, with a memory buffer of the maximum size of 2000, and we randomly remove a batch of stored samples (batch size is 10) when the memory is full. Then we estimate ${\mathcal{R}}(h,{\mathcal{P}}_i)$ (target risk on all visited training samples), ${\mathcal{R}}(h,h_{{\mathcal{M}}_i},{\mathbb{P}}_{{\mathcal{M}}_i})$ (source risk on the memory). We plot the results in Fig.~\ref{riskResult2}-a, where ``Random (Source risk)'' represents the source risk of a single classifier. The results show that the source risk always keeps stable, and the target risk is small for a few initial training steps since the memory can capture all information of visited samples. However, as the number of training steps grows, the target risk is increased, which is caused by the memory that loses previous samples, theoretically explained in Theorem 1. We also evaluate the risk of the baseline on all testing sets (target risk) and plot the results in Fig.~\ref{riskResult2}-b where it is observed that the single model always leads to a large target risk during the training.

We evaluate the source and target risks for the proposed ODDL under Split MNIST and plot the results in Fig.~\ref{riskResult2}. The performance of ODDL on the distribution ${\mathcal{P}}_i$ (Target risk)  does not degenerate during the whole training phase. At the same time, the baseline tends to increase the target risk on ${\mathcal{P}}_i$ as the training steps increase, as shown in Fig.~\ref{riskResult2}-a. Finally, the risk on all target distributions (all categories in the testing set) from each training step is shown in Fig.~\ref{riskResult2}-b, where the proposed ODDL minimizes the target risk as gaining more knowledge during the training. In contrast, the baseline invariably leads to a large target risk even when the number of training steps increases. These results show that ODDL can relieve forgetting and achieve better generalization than the random approach on all target sets.

\vspace{-5pt}
\subsection{Results on TFCL benchmark}

We provide the results in Tab.~\ref{classification} where * and $\dag$ denote the results cited from \cite{ContinualPrototype} and \cite{GradientTFCL}, respectively. We compare with several baselines including: finetune that directly trains a classifier on the data stream, GSS \cite{GradientLifelong}, Dynamic-OCM \cite{OCM}, MIR \cite{OnlineContinualLearning}, Gradient Episodic Memory (GEM) \cite{GradientEpisodic}, Incremental Classifier and Representation Learning (iCARL) \cite{icarl}, Reservoir \cite{reservoir}, CURL, CNDPM, CoPE, ER + GMED and ER$_a$ + GMED \cite{GradientTFCL} where ER is the 
Experience Replay \cite{ExperienceReplay} and ER$_a$ is ER with data augmentation. The number of components in ODDL-S and ODDL for Split MNIST, Split CIFAR10 and Split CIFAR100 is 7, 9, 7, respectively. The proposed approach outperforms CNDPM, which uses more parameters, on all datasets and achieves state-of-the-art performance.

\begin{wraptable}{r}{5.5cm}
\centering
\vspace{-16pt}
	\caption{Classification accuracy for 20 runs when testing various models on Split MImageNet and Permuted MNIST.}
	\vspace{7pt}
\renewcommand\arraystretch{1.15}
\scriptsize
\setlength{\tabcolsep}{1.2mm}{\begin{tabular}{@{}l cc c c@{} } 
\toprule
\textBF{Methods}   &\textBF{Split MImageNet} & \textBF{Permuted
MNIST}
  \\
  \midrule 
 ER$_{a}$ &25.92 $\pm$ 1.2 & 78.11 $\pm$ 0.7 \\
 ER + GMED &27.27 $\pm$ 1.8 & 78.86 $\pm$ 0.7 \\
 MIR+GMED &26.50 $\pm$ 1.3 & 79.25 $\pm$ 0.8 \\
  MIR&25.21 $\pm$ 2.2&79.13 $\pm$ 0.7 \\
  CNDPM&27.12 $\pm$1.5&80.68 $\pm$ 0.7
  \\
	 \hline
	 \hline
	   ODDL & 27.45 $\pm$ 0.9&82.33 $\pm$ 0.6 \\
	   ODDL-S & \textBF{28.68} $\pm$ 1.5&\textBF{83.56} $\pm$ 0.5   \\
\bottomrule 
\end{tabular}
\label{MINIImageNet_tab}
}
\end{wraptable}

In the following, we also evaluate the performance of the proposed approach on the large-scale dataset (MINI-ImageNet \cite{TinyImageNet}), and Permuted MNIST \cite{empiricalCL}. We split MINI-ImageNet into 20 tasks (See details in Appendix-F1 from SM), namely Split MImageNet. We follow the setting from \cite{OnlineContinualLearning} where the maximum memory size is $10K$, and a smaller version of ResNet-18 \cite{DeepRes} is used as the classifier. The hyperparameter $\lambda$ used for learning Split MINI-ImageNet and Permuted MNIST is equal to 1.2 and 1.5, respectively. We compare with several state-of-the-art methods under Split MImageNet, reported in Tab.~\ref{MINIImageNet_tab}, where the results of other baselines are cited from \cite{GradientTFCL}. These results show that the proposed ODDL-S outperforms different baselines under the challenging dataset. 

\vspace{-5pt}
\subsection{Ablation study}
\label{sec:ablation} 

We investigate whether the proposed discrepancy-based criterion can provide better signals for the expansion of ODDL-S. We train ODDL-S on Split MNIST, where we record the variance of tasks and the number of components in each training step. We plot the results in 
Fig.~\ref{riskResult2}-c where ``task'' represents the number of tasks in each training step. We observe that the proposed discrepancy-based criterion can detect the data distribution shift accurately, allowing ODDL-S to expand the network architecture each time when detecting the data distribution shift. This also encourages the proposed ODDL-S to use a minimal number of components while achieving optimal performance, as discussed in Lemma 2. We also provide the analysis of how to maximize the memory bound in Appendix-F2.2, while more ablation study results are provided in Appendix-F.2 from the supplementary material. 

\vspace{-5pt}
\section{Conclusion}
\vspace{-5pt}
In this paper, we develop a novel theoretical framework for Task-Free Continual Learning (TFCL), by defining a statistical discrepancy distance. Inspired by the theoretical analysis, we propose the Online Discrepancy Distance Learning enabled by a memory buffer sampling (ODDL-S) model, which trades off between the model's complexity and performance. The memory buffer sampling mechanism ensures the information diversity learning.
The proposed theoretical analysis provides new insights into the model's forgetting behaviour during each training step of TFCL.  Experimental results on several TFCL benchmarks show that the proposed ODDL-S achieves state-of-the-art performance.

\bibliographystyle{plain}
\bibliography{main}

\begin{thebibliography}{10}

\bibitem{GradientLifelong}
R.~Aljundi, M.~Lin, B.~Goujaud, and Y.~Bengio.
\newblock Gradient based sample selection for online continual learning.
\newblock In {\em Advances in Neural Information Processing Systems (NeurIPS),
  arXiv preprint arXiv:1903.08671}, 2019.

\bibitem{OnlineContinualLearning}
Rahaf Aljundi, Eugene Belilovsky, Tinne Tuytelaars, Laurent Charlin, Massimo
  Caccia, Min Lin, and Lucas Page-Caccia.
\newblock Online continual learning with maximal interfered retrieval.
\newblock In {\em Advances in Neural Information Processing Systems (NeurIPS),
  arXiv preprint arXiv:1908.04742}, 2019.

\bibitem{taskFree_CL}
Rahaf Aljundi, Klaas Kelchtermans, and Tinne Tuytelaars.
\newblock Task-free continual learning.
\newblock In {\em Proc. of IEEE/CVF Conf. on Computer Vision and Pattern
  Recognition}, pages 11254--11263, 2019.

\bibitem{minMaxLearning}
Ver{\'{o}}nica {\'{A}}lvarez, Santiago Mazuelas, and Jos{\'{e}}~Antonio Lozano.
\newblock Minimax classification under concept drift with multidimensional
  adaptation and performance guarantees.
\newblock In {\em Proc. International Conference on Machine Learning (ICML),
  vol. PMLR 162}, pages 486--499, 2022.

\bibitem{RainbowMemory}
Jihwan Bang, Heesu Kim, YoungJoon Yoo, Jung-Woo Ha, and Jonghyun Choi.
\newblock Rainbow memory: Continual learning with a memory of diverse samples.
\newblock In {\em Proc. of IEEE/CVF Conf. on Computer Vision and Pattern
  Recognition (CVPR)}, pages 8218--8227, 2021.

\bibitem{TinyLifelong}
A.~Chaudhry, M.~Rohrbach, M.~Elhoseiny, T.~Ajanthan, P.~Dokania, P.~H.~S. Torr,
  and M.'A. Ranzato.
\newblock On tiny episodic memories in continual learning.
\newblock {\em arXiv preprint arXiv:1902.10486}, 2019.

\bibitem{BoostingTransfer}
W.~Dai, Q.~Yang, G.~R. Xue, and Y.~Yu.
\newblock Boosting for transfer learning.
\newblock In {\em Proc. Int Conf. on Machine Learning (ICML)}, pages 193--200,
  2007.

\bibitem{ContinualPrototype}
Matthias De~Lange and Tinne Tuytelaars.
\newblock Continual prototype evolution: Learning online from non-stationary
  data streams.
\newblock In {\em Proc. of the IEEE/CVF Int. Conf. on Computer Vision (ICCV)},
  pages 8250--8259, 2021.

\bibitem{empiricalCL}
Ian~J Goodfellow, Mehdi Mirza, Da~Xiao, Aaron Courville, and Yoshua Bengio.
\newblock An empirical investigation of catastrophic forgetting in
  gradient-based neural networks.
\newblock In {\em arXiv preprint arXiv:1312.6211}, 2014.

\bibitem{DeepRes}
K.~He, X.~Zhang, S.~Ren, and J.~Sun.
\newblock Deep residual learning for image recognition.
\newblock In {\em Proc. of IEEE Conf. on Computer Vision and Pattern
  Recognition (CVPR)}, pages 770--778, 2016.

\bibitem{Distilling_nets}
G.~Hinton, O.~Vinyals, and J.~Dean.
\newblock Distilling the knowledge in a neural network.
\newblock In {\em Proc. NIPS Deep Learning Workshop, arXiv preprint
  arXiv:1503.02531}, 2014.

\bibitem{GradientTFCL}
Xisen Jin, Arka Sadhu, Junyi Du, and Xiang Ren.
\newblock Gradient-based editing of memory examples for online task-free
  continual learning.
\newblock In {\em Advances in Neural Information Processing Systems (NeurIPS),
  arXiv preprint arXiv:2006.15294}, 2021.

\bibitem{LessForgetting}
H.~Jung, J.~Ju, M.~Jung, and J.~Kim.
\newblock Less-forgetting learning in deep neural networks.
\newblock In {\em Proc. AAAI Conf. on Artificial Intelligence}, volume~32,
  pages 3358--3365, 2018.

\bibitem{AchievingForgetting}
Zixuan Ke, Bing Liu, Nianzu Ma, Hu~Xu, and Lei Shu.
\newblock Achieving forgetting prevention and knowledge transfer in continual
  learning.
\newblock {\em Advances in Neural Information Processing Systems}, 34, 2021.

\bibitem{VAE}
D.~P. Kingma and M.~Welling.
\newblock Auto-encoding variational {B}ayes.
\newblock {\em arXiv preprint arXiv:1312.6114}, 2013.

\bibitem{EWC}
J.~Kirkpatrick, R.~Pascanu, N.~Rabinowitz, J.~Veness, G.~Desjardins, A.~A.
  Rusu, K.~Milan, J.~Quan, T.~Ramalho, A.~Grabska-Barwinska, D.~Hassabis,
  C.~Clopath, D.~Kumaran, and R.~Hadsell.
\newblock Overcoming catastrophic forgetting in neural networks.
\newblock {\em Proc. of the National Academy of Sciences (PNAS)},
  114(13):3521--3526, 2017.

\bibitem{OptimalCL}
Jeremias Knoblauch, Hisham Husain, and Tom Diethe.
\newblock Optimal continual learning has perfect memory and is {NP}-hard.
\newblock In {\em Proc. Int. Conf. on Machine Learning (ICML), vol PMLR 119},
  pages 5327--5337, 2020.

\bibitem{CIFAR10}
Alex Krizhevsky and Geoffrey Hinton.
\newblock Learning multiple layers of features from tiny images.
\newblock Technical report, Univ. of Toronto, 2009.

\bibitem{CL_Bayesian}
Richard Kurle, Botond Cseke, Alexej Klushyn, Patrick van~der Smagt, and Stephan
  Günnemann.
\newblock Continual learning with {Bayesian} neural networks for non-stationary
  data.
\newblock In {\em Int. Conf. on Learning Representations (ICLR)}, 2020.

\bibitem{AutonomousCar}
Sampo Kuutti, Richard Bowden, Yaochu Jin, Phil Barber, and Saber Fallah.
\newblock A survey of deep learning applications to autonomous vehicle control.
\newblock {\em IEEE Transactions on Intelligent Transportation Systems},
  22(2):712--733, 2020.

\bibitem{TinyImageNet}
Ya~Le and Xuan Yang.
\newblock Tiny image{Net} visual recognition challenge.
\newblock Technical report, Univ. of Stanford, 2015.

\bibitem{MNIST}
Y.~LeCun, L.~Bottou, Y.~Bengio, and P.~Haffner.
\newblock Gradient-based learning applied to document recognition.
\newblock {\em Proc. of the IEEE}, 86(11):2278--2324, 1998.

\bibitem{CLTeacherStudent}
Sebastian Lee, Sebastian Goldt, and Andrew Saxe.
\newblock Continual learning in the teacher-student setup: Impact of task
  similarity.
\newblock In {\em Proc. Int. Conf. on Machine Learning (ICML), vol. PMLR 139},
  pages 6109--6119, 2021.

\bibitem{NeuralDirichelt}
Soochan Lee, Junsoo Ha, Dongsu Zhang, and Gunhee Kim.
\newblock A neural {D}irichlet process mixture model for task-free continual
  learning.
\newblock In {\em Int. Conf. on Learning Representations (ICLR), arXiv preprint
  arXiv:2001.00689}, 2020.

\bibitem{Lwf}
Z.~Li and D.~Hoiem.
\newblock Learning without forgetting.
\newblock {\em IEEE Trans. on Pattern Analysis and Machine Intelligence},
  40(12):2935--2947, 2017.

\bibitem{TRGP}
Sen Lin, Li~Yang, Deliang Fan, and Junshan Zhang.
\newblock {TRGP}: Trust region gradient projection for continual learning.
\newblock In {\em Int. Conf. on Learning Representations (ICLR), arXiv preprint
  arXiv:2202.02931}, 2022.

\bibitem{GradientEpisodic}
David Lopez-Paz and Marc'Aurelio Ranzato.
\newblock Gradient episodic memory for continual learning.
\newblock In {\em Advances in Neural Information Processing Systems}, pages
  6467--6476, 2017.

\bibitem{RepresentationalContinuity}
Divyam Madaan, Jaehong Yoon, Yuanchun Li, Yunxin Liu, and Sung~Ju Hwang.
\newblock Representational continuity for unsupervised continual learning.
\newblock In {\em Int. Conf. on Learning Representations (ICLR), arXiv preprint
  arXiv:2110.06976}, 2022.

\bibitem{domainTheory}
Yishay Mansour, Mehryar Mohri, and Afshin Rostamizadeh.
\newblock Domain adaptation: Learning bounds and algorithms.
\newblock In {\em Proc. Conf. on Learning Theory (COLT), arXiv preprint
  arXiv:2002.06715}, 2009.

\bibitem{Theory_drifiting}
Mehryar Mohri and Andres~Munoz Medina.
\newblock New analysis and algorithm for learning with drifting distributions.
\newblock In {\em International Conference on Algorithmic Learning Theory},
  pages 124--138. Springer, 2012.

\bibitem{VCL}
Cuong~V Nguyen, Yingzhen Li, Thang~D Bui, and Richard~E Turner.
\newblock Variational continual learning.
\newblock In {\em Int. Conf. on Learning Representations (ICLR), arXiv preprint
  arXiv:1710.10628}, 2018.

\bibitem{LifeLong_review}
G.~I. Parisi, R.~Kemker, J.~L. Part, C.~Kanan, and S.~Wermter.
\newblock Continual lifelong learning with neural networks: A review.
\newblock {\em Neural Networks}, 113:54--71, 2019.

\bibitem{LifelongIID}
Anastasia Pentina and Christoph~H Lampert.
\newblock Lifelong learning with non-iid tasks.
\newblock In {\em Proc. Advances in Neural Information Processing Systems
  (NIPS)}, pages 1540--1548, 2015.

\bibitem{LearnAdd}
R.~Polikar, L.~Upda, S.~S. Upda, and Vasant Honavar.
\newblock Learn++: An incremental learning algorithm for supervised neural
  networks.
\newblock {\em IEEE Trans. on Systems Man and Cybernetics, Part C},
  31(4):497--508, 2001.

\bibitem{LookingBack}
Mozhgan PourKeshavarzi, Guoying Zhao, and Mohammad Sabokrou.
\newblock Looking back on learned experiences for class/task incremental
  learning.
\newblock In {\em Int. Conf. on Learning Representations (ICLR)}, 2022.

\bibitem{Gdumb}
Ameya Prabhu, Philip~HS Torr, and Puneet~K Dokania.
\newblock {GDumb}: A simple approach that questions our progress in continual
  learning.
\newblock In {\em Proc. European Conference on Computer Vision (ECCV), vol.
  LNCS 12347}, pages 524--540, 2020.

\bibitem{CL_TradeOff}
Krishnan Raghavan and Prasanna Balaprakash.
\newblock Formalizing the generalization-forgetting trade-off in continual
  learning.
\newblock {\em Advances in Neural Information Processing Systems}, 34, 2021.

\bibitem{LifelongUnsupervisedVAE}
Dushyant Rao, Francesco Visin, Andrei~A. Rusu, Yee~Whye Teh, Razvan Pascanu,
  and Raia Hadsell.
\newblock Continual unsupervised representation learning.
\newblock In {\em Proc. Neural Inf. Proc. Systems (NIPS)}, pages 7645--7655,
  2019.

\bibitem{icarl}
Sylvestre-Alvise Rebuffi, Alexander Kolesnikov, Georg Sperl, and Christoph~H
  Lampert.
\newblock {iCaRL}: Incremental classifier and representation learning.
\newblock In {\em Proc. of the IEEE Conf. on Computer Vision and Pattern
  Recognition (CVPR)}, pages 2001--2010, 2017.

\bibitem{LifeLong_combination}
B.~Ren, H.~Wang, J.~Li, and H.~Gao.
\newblock Life-long learning based on dynamic combination model.
\newblock {\em Applied Soft Computing}, 56:398--404, 2017.

\bibitem{OnlineStructuredLaplace}
Hippolyt Ritter, Aleksandar Botev, and David Barber.
\newblock Online structured {L}aplace approximations for overcoming
  catastrophic forgetting.
\newblock In {\em Advances in Neural Information Processing Systems (NeurIPS)},
  volume~31, pages 3742--3752, 2018.

\bibitem{ExperienceReplay}
David Rolnick, Arun Ahuja, Jonathan Schwarz, Timothy~P. Lillicrap, and Gregory
  Wayne.
\newblock Experience replay for continual learning.
\newblock In {\em Advances in Neural Information Processing Systems 34
  (NeurIPS)}, pages 348--358, 2019.

\bibitem{Generative_replay}
H.~Shin, J.~K. Lee, J.~Kim, and J.~Kim.
\newblock Continual learning with deep generative replay.
\newblock In {\em Advances in Neural Inf. Proc. Systems (NIPS)}, pages
  2990--2999, 2017.

\bibitem{InfomrationCL}
Shengyang Sun, Daniele Calandriello, Huiyi Hu, Ang Li, and Michalis Titsias.
\newblock Information-theoretic online memory selection for continual learning.
\newblock In {\em Int. Conf. on Learning Representations (ICLR), arXiv preprint
  arXiv:2204.04763}, 2022.

\bibitem{reservoir}
Jeffrey~S Vitter.
\newblock Random sampling with a reservoir.
\newblock {\em ACM Transactions on Mathematical Software (TOMS)}, 11(1):37--57,
  1985.

\bibitem{LifelongVAEGAN}
Fei Ye and Adrian~G. Bors.
\newblock Learning latent representations across multiple data domains using
  lifelong {VAEGAN}.
\newblock In {\em Proc. European Conf. on Computer Vision (ECCV), vol. LNCS
  12365}, pages 777--795, 2020.

\bibitem{Lifelonginterpretable}
Fei Ye and Adrian~G. Bors.
\newblock Lifelong learning of interpretable image representations.
\newblock In {\em Proc. Int. Conf. on Image Processing Theory, Tools and
  Applications (IPTA)}, pages 1--6, 2020.

\bibitem{MixtureOfVAEs}
Fei Ye and Adrian~G Bors.
\newblock Mixtures of variational autoencoders.
\newblock In {\em Proc. Int. Conf. on Image Processing Theory, Tools and
  Applications (IPTA)}, pages 1--6, 2020.

\bibitem{InfoVAEGAN_conference}
Fei Ye and Adrian~G. Bors.
\newblock {InfoVAEGAN}: Learning joint interpretable representations by
  information maximization and maximum likelihood.
\newblock In {\em Proc. IEEE Int. Conf. on Image Processing (ICIP)}, pages
  749--753, 2021.

\bibitem{JontLatentVAEs}
Fei Ye and Adrian~G Bors.
\newblock Learning joint latent representations based on information
  maximization.
\newblock {\em Information Sciences}, 567:216--236, 2021.

\bibitem{LifelongInfinite}
Fei Ye and Adrian~G. Bors.
\newblock Lifelong infinite mixture model based on knowledge-driven {D}irichlet
  process.
\newblock In {\em Proc. of the IEEE Int. Conf. on Computer Vision (ICCV)},
  pages 10695--10704, 2021.

\bibitem{LifelongMixuteOfVAEs}
Fei Ye and Adrian~G. Bors.
\newblock Lifelong mixture of variational autoencoders.
\newblock {\em IEEE Transactions on Neural Networks and Learning Systems},
  pages 1--14, 2021.

\bibitem{LifelongTwin}
Fei Ye and Adrian~G. Bors.
\newblock Lifelong twin generative adversarial networks.
\newblock In {\em Proc. IEEE Int. Conf. on Image Processing (ICIP)}, pages
  1289--1293, 2021.

\bibitem{OCM}
Fei Ye and Adrian~G Bors.
\newblock Continual variational autoencoder learning via online cooperative
  memorization.
\newblock {\em arXiv preprint arXiv:2207.10131}, 2022.

\bibitem{DeepMixtureVAE}
Fei Ye and Adrian~G. Bors.
\newblock Deep mixture generative autoencoders.
\newblock {\em IEEE Transactions on Neural Networks and Learning Systems},
  33(10):5789--5803, 2022.

\bibitem{LifelongEvolved}
Fei Ye and Adrian~G Bors.
\newblock Learning an evolved mixture model for task-free continual learning.
\newblock {\em arXiv preprint arXiv:2207.05080}, 2022.

\bibitem{LifelongGraph}
Fei Ye and Adrian~G. Bors.
\newblock Lifelong generative modelling using dynamic expansion graph model.
\newblock {\em Proceedings of the AAAI Conference on Artificial Intelligence},
  36(8):8857--8865, Jun. 2022.

\bibitem{LifelongTeacherStudent}
Fei Ye and Adrian~G. Bors.
\newblock Lifelong teacher-student network learning.
\newblock {\em IEEE Transactions on Pattern Analysis and Machine Intelligence},
  44(10):6280--6296, 2022.

\bibitem{OnlineCoreset}
Jaehong Yoon, Divyam Madaan, Eunho Yang, and Sung~Ju Hwang.
\newblock Online coreset selection for rehearsal-based continual learning.
\newblock In {\em Int. Conf. on Learning Representations (ICLR), arXiv preprint
  arXiv:2106.01085}, 2022.

\bibitem{BridgingTheory}
Yuchen Zhang, Tianle Liu, Mingsheng Long, and Michael Jordan.
\newblock Bridging theory and algorithm for domain adaptation.
\newblock In {\em Proc. International Conference on Machine Learning (ICML),
  vol. PMLR 97}, pages 7404--7413, 2019.

\bibitem{InfoVAE}
Shengjia Zhao, Jiaming Song, and Stefano Ermon.
\newblock Info{VAE}: Balancing learning and inference in variational
  autoencoders.
\newblock In {\em Proc. AAAI Conf. on Artif. Intel.}, volume~33, pages
  5885--5892, 2019.

\end{thebibliography}

\end{document}